\documentclass[dvipsnames,format=sigconf,anonymous=false,review=false]{acmart}

\AtBeginDocument{%
  \providecommand\BibTeX{{%
    \normalfont B\kern-0.5em{\scshape i\kern-0.25em b}\kern-0.8em\TeX}}}

\acmDOI{10.1145/3638529.3654030}
\acmISBN{979-8-4007-0494-9/24/07}
\acmConference[GECCO '24]{The Genetic and Evolutionary Computation Conference 2024}{July 14--18, 2024}{Melbourne, Australia}
\acmYear{2024}
\copyrightyear{2024}

\begin{document}

\title[RETRACTED Evolving Form and Function]{[RETRACTED]Evolving Form and Function: Dual-Objective Optimization in Neural Symbolic Regression Networks}

\author{Amanda Bertschinger}
\email{ambertsc@uvm.edu}
\orcid{1234-5678-9012}
\affiliation{%
  \institution{University of Vermont}
  \city{South Burlington}
  \state{Vermont}
  \country{USA}
  \postcode{05403}
}

\author{James Bagrow}
\affiliation{%
    \institution{University of Vermont}
    \city{South Burlington}
    \state{Vermont}
    \country{USA}
    \postcode{05403}}

\author{Joshua Bongard}
\affiliation{%
  \institution{University of Vermont}
    \city{South Burlington}
  \state{Vermont}
  \country{USA}
  \postcode{05403}
}
\renewcommand{\shortauthors}{Bertschinger, et al.}

\begin{abstract}
Data increasingly abounds, but distilling their underlying relationships down to something interpretable remains challenging. One approach is genetic programming, which `symbolically regresses' a data set down into an equation.
However, symbolic regression (SR) faces the issue of requiring training from scratch for each new dataset. To generalize across all datasets, deep learning techniques have been applied to SR.
These networks, however, are only able to be trained using a symbolic objective: NN-generated and target equations are symbolically compared. But this does not consider the predictive power of these equations, which could be measured by a behavioral objective that compares the generated equation's predictions to actual data.
Here we introduce a method that combines gradient descent and evolutionary computation to yield neural networks that minimize the symbolic and behavioral errors of the equations they generate from data. 
As a result, these evolved networks are shown to generate more symbolically and behaviorally accurate equations than those generated by networks trained by state-of-the-art gradient based neural symbolic regression methods.
We hope this method suggests that evolutionary algorithms, combined with gradient descent, can improve SR results by yielding equations with more accurate form and function.
\end{abstract}



\begin{teaserfigure}
  \includegraphics[width=\textwidth]{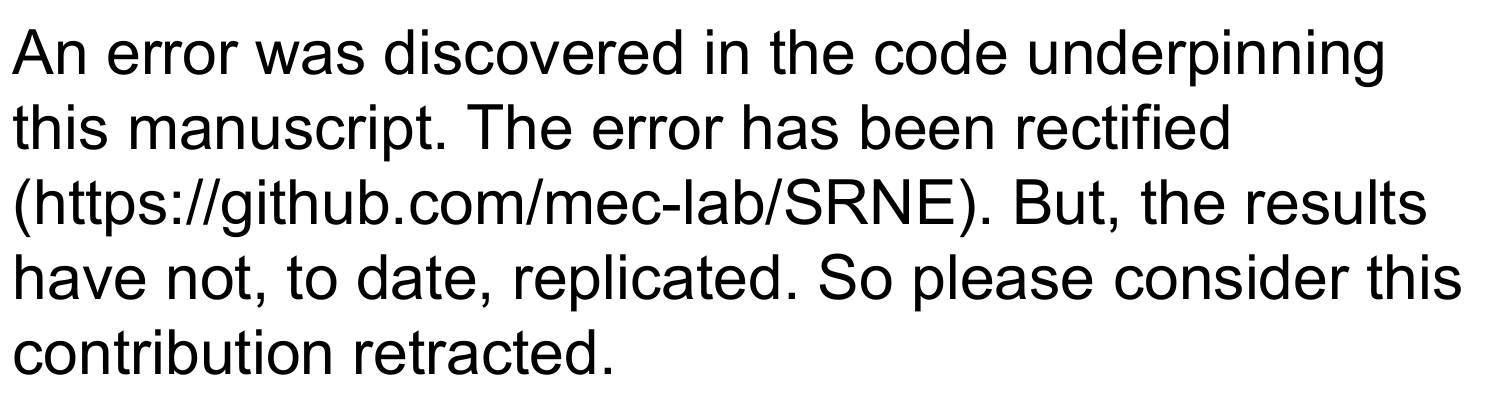}
\end{teaserfigure}
\begin{teaserfigure}
  \includegraphics[width=\textwidth]{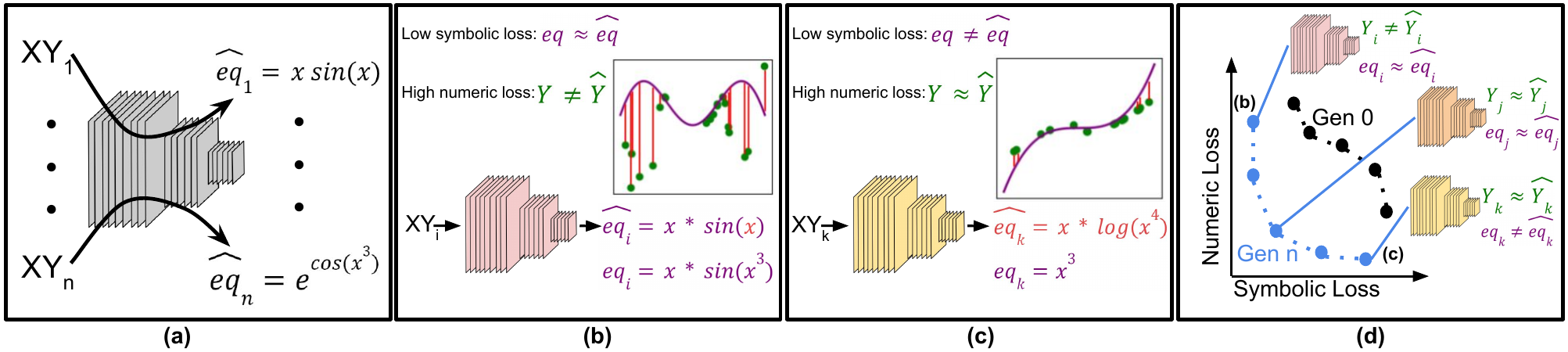}
  \vspace{-0.4cm}
  \caption{The Symbolic Regression NeuroEvolution (SRNE) method. (a) Generalized neural symbolic regression (GNSR) networks are trained to take any XY data as input and output an equation that describes said data. (b) Some networks have low symbolic loss (the difference between a generated equation's form ($\widehat{eq}$) and that of the hidden equation ($eq$) underlying their training data (XY)) but high numeric loss (the difference between a generated equation's predictions ($\hat{Y}$) and a hidden equation's values ($Y$)). (c) Others have high symbolic loss but low numeric loss. (d) SRNE evolves populations of data-to-equation neural networks to reduce their symbolic loss and numeric loss simultaneously to create the optimal case where networks have low symbolic and numeric loss. }
  \label{fig:teaser}
\end{teaserfigure}

\maketitle

\section{Introduction}
With the overwhelming large amount of data being collected worldwide, the ability to process and understand this data is paramount. However, raw data is generally very difficult for people to understand without the intervention of experts. These experts are generally tasked with describing the underlying relationships in the data, and often this requires finding equations that describe the dataset. Even experts may struggle with this difficult but important task. In order to help compensate for this, techniques such as regression are used to help infer these equations.

Classical regression fits a predetermined "skeleton" equation to a dataset using coefficients~\cite{Motulsky1987,Weisberg2005}. However, this approach still suffers from the need for expert intervention in picking the functional form of an equation to fit. To compensate for this weakness and fully automate the creation of an equation to describe a given dataset, symbolic regression (SR) can be used. SR traditionally uses genetic programming (GPSR) in order to build an equation from a corpus of mathematical operators~\cite{Koza1992,Koza1994}. Using GPSR, equation trees are evolved using an evolutionary algorithm to fit a given dataset more and more accurately. It is an approach that can give very accurate equations, and as such GPSR has been well studied~\cite{Keijzer2003,Keijzer2004,Schmidt2008,Schmidt2009,Arnaldo2014,Virgolin2021}.

GPSR does have disadvantages though, primarily in the large amount of resources it requires: for each new dataset, GPSR needs to be trained from scratch. One recently proposed solution to this is neural symbolic regression (NSR), in which networks transform data into equations. Two primary types of neural symbolic regression exist: deep neural symbolic regression (DSR)~\cite{Martius2016,Sahoo2018,Kim2020,Costa2020,Petersen2019,Zhang2023,Boddupalli2023} and generalized neural symbolic regression (GNSR)~\cite{Biggio2020,Biggio2021,Valipour2021,Kamienny2022,Vastl2022,Li2022,Bendinelli2023}. DSR methods train a network to generate one equation from one data set by utilizing neural networks' structure and their activation functions to represent an equation. DSR is more computationally efficient than GPSR, but suffers the same drawback of requiring \textit{de novo} retraining per new dataset that GPSR does. GNSR, however, leverages the scale of deep learning to pretrain a neural network on many unique dataset-equation pairs, and then uses transfer learning to allow the network to predict new equations from previously unseen datasets. Thus, though there is a large upfront cost to training a GNSR network, once it is trained it can quickly predict an equation for any given dataset.

GNSR shows promise for overcoming the problems of classical regression and traditional SR techniques. However, it is known that the methods are often over-reliant on coefficient fitting after the networks generate an equation~\cite{Bertschinger2023}: GNSR methods generate coefficient free equations, and a post hoc coefficient fitting step allows them to `cheat' by generating symbolically inaccurate equations. When deprived of access to coefficient fitting, GNSR methods often yield equations that are still symbolically accurate (the form of generated equations are similar to that of the hidden equations used to generate the data they were trained on) or numerically accurate (the predictions of generated equations and hidden equations are similar), but not both. 

The difference between symbolic and numeric measures of equation accuracy is an important one: a symbolic loss metric evaluates a generated equation's syntactic accuracy whereas a numeric loss metric evaluates a generated equation's semantic accuracy. Importantly, an equation with low symbolic loss does not guarantee it will enjoy low numeric loss (Fig. \ref{fig:teaser}(b)) and vice versa (Fig. \ref{fig:teaser}(d)). One current problem with GNSR that is contributing to the observed inaccuracy across both kinds metrics is that all seven of the current GNSR methods~\cite{Biggio2020,Biggio2021,Valipour2021,Kamienny2022,Vastl2022,Li2022,Bendinelli2023} are trained only on symbolic loss. Both symbolic and numeric metrics reveal orthogonal information about an equation's accuracy~\cite{Bertschinger2023}, and so GNSR methods need to be trained against both. However, with current techniques this has not been possible, for two reasons. First, it is challenging to backpropagate loss using two loss functions. Second, generated equations are currently numerically evaluated in a non-differentiable way. This means that their numeric loss can only be calculated during testing time, and cannot yet be employed to backpropagate numeric loss.

In order to overcome this, here we present a GNSR method where the weights of networks are evolved rather than trained through backpropagation. This neuroevolution allows our networks to, for the first time, be jointly optimized on numeric and symbolic loss. We show that this multi-objective optimization leads to networks that are able to predict equations with better symbolic and numeric loss than existing methods.

\section{Related Works}
Symbolic regression, particularly GPSR\cite{Koza1992}, has been studied for decades now. Improvements to the algorithms used have been made since, and continue to be an active area of research today~\cite{LaCava2021,Zojaji2022,Haider2023,Yang2023,Fleck2024}. DSR is a newer sub-field of SR, first appearing in the mid-2010's~\cite{Martius2016}. It is just as active an area of research as GPSR, with new methods for DSR networks being studied~\cite{Petersen2019,Zhang2023,Boddupalli2023}.

The newest sub-field of SR is GNSR, with 7 reported methods in the literature to date~\cite{Biggio2020,Biggio2021,Valipour2021,Kamienny2022,Vastl2022,Li2022,Bendinelli2023}. Inspired by the success of transformer architectures in areas such as large language models~\cite{Vaswani2017,Keskar2019,Wang2019language}, most of these methods~\cite{Biggio2021,Valipour2021,Kamienny2022,Vastl2022,Li2022,Bendinelli2023} use transformer networks. These networks are all data-to-equation networks, where numeric data is fed into the transformer which then outputs a tokenized equation (Fig. \ref{fig:teaser}(a)).
However, they all use only one loss metric during training. Multi-objective loss functions are traditionally difficult for gradient-based methods to handle. However, evolutionary methods admit them easily in multi-objective optimization frameworks. It is for this reason that we here introduce and investigate a evolutionary, multi-objective approach to GNSR: data-to-equation models are evolved against symbolic and numeric loss objectives.

Using evolutionary computation to optimize neural networks is a common practice\cite{Jin2009,Feurer2019,Biswas2021,Tani2021}. Generally, this takes the form of hyper-parameter optimization or even optimization of neural network structure. However, the neuroevolution of deep learning networks' weights has also been broadly studied in literature for many years\cite{Montana1989,Rocha2007,Floreano2008,Ding2013,Stanley2019,Galvan2021}. One of the most common evolutionary algorithms for neuroevolution, NEAT~\cite{Stanley2002} uses evolutionary methods to optimize both structure and weights of neural networks. Here, however, we use traditional neuroevolution where only the weights are evolved.

\section{Methods}
Our method, which we call symbolic regression neuroevolution (SRNE), trains and evolves neural networks to solve the generalized data-to-equation problem: that is, learning how to immediately compress any new data set ($XY$) into a good formal description ($eq$) of it, without requiring further training or evolution. To do so, we evolve the synaptic weights of a GNSR network (Fig. \ref{fig:teaser}(a)). The network itself is being evolved in our method; the equations produced by the network are simply used to define how well our networks perform at their given task in order to allow for evolutionary selection. We here define goodness as a multi-objective measure: a good equation should have good form and good function. We define good form as the ability of a generated equation to match that likely to be derived by humans for the same data set. We define good function as the ability of an equation to predict unseen data from the data set it was generated from.
A visual summary of our method is provided in Fig. \ref{fig:teaser}.
We next describe the architecture of our AI models    (sect. \ref{modelArch}),
then how they are pre-trained    (sect. \ref{pre-training}), 
evolved                          (sect. \ref{evolution}), 
validated after evolution ceases (sect. \ref{testing}), as well as 
the state-of-the-art GNSR networks we compare against (sect. \ref{comparison}).
Code for our method can be accessed at the SRNE git repo (https://github.com/mec-lab/SRNE).

\subsection{Network Architecture}
\label{modelArch}
In this section, the architecture of the transformer models used in our populations, as well as the way training and testing equations equations are represented and generated, is discussed.

\subsubsection{Network Architecture}
For the networks that make up the individuals in our population, we use a modified version of Valipour's~\cite{Valipour2021} transformer network. It is comprised of two networks: a convolutional PointNet~\cite{Qi2017} to encode the input $XY$ data into an order-invariant embedding that feeds into the main transformer~\cite{Vaswani2017} network which predicts the equations.

The PointNet architecture has convolutional and fully connected layers. We remove the standard batch normalization layers in order to allow for the network to have better access to the dependency between $X$ and $y$ values, especially the more extreme $y$ values produced by some equations.

The transformer network is a standard GPT~\cite{Radford2018} network with 8 transformer blocks. Each block has multi-headed attention and an MLP with two fully-connected layers, Gaussian Error Linear Units~\cite{Hendrycks2016}, dropout, and a final fully connected layer. In all the transformer blocks we remove the normalization layers that were present in Valipour's model.

\subsubsection{Equation representation.}
To produce the training equations that the model is first pretrained and then evolved on, we randomly generate equations from the following set of mathematical operators: $[\sin,\cos,\texttt{pow},+,*,\exp,\log]$. These operators were chosen to allow for a variety of equations, without introducing the potential for invalid equations that operators like division may result in (in regards to divide-by-zero errors). Equations are represented as an ordered list of primitives (mathematical operators, variables, and constants for the $pow$ operator) which corresponds to a pre-order traversal of the equation's tree representation. The equations are constrained to a single independent variable $x$ and a maximum of 30 primitives per equation, though generally the generated equations have a much lower number of primitives. The $pow$ operator is constrained to constants between 2 and 4 inclusive, but can access variables freely. No coefficients outside of 1 are allowed for any terms in the equations. Once valid equations have been randomly generated from this operator set, the equations are tokenized. Each unique primitive is assigned a unique integer ID and the ordered primitive list is transformed into an ordered integer list by a primitive-to-token dictionary. The networks in our population are then trained to produce these tokenized equations from a corresponding set of $X,Y$ data generated by $y_i = eq(x_i)$ for $i = 1,...,30$.

\subsection{Pre-training.}
\label{pre-training}
In this section, the details of the initial pre-training of the networks in our starting population is discussed.

All networks in the initial parent population undergo an initial pre-training of their weights using gradient descent techniques. To achieve this, 25 networks are trained on a dataset of 5000 randomly generated data-equation pairs. The details and constraints of these data-equations pairs and their generation are discussed in sect. \ref{modelArch}. The loss metric used in this initial pre-training is symbolic cross-entropy (CE) loss (definition given below) only. The model weights from 25 trials are saved and shared across the trials of the evolutionary algorithm. An initial population for evolution are randomly assigned the pre-trained weights from one of these models.

\subsection{Evolution.}
\label{evolution}
In this section we discuss the details of our evolutionary algorithm, including evaluation of the population members, selection of population members to survive between generations, and details of how parents of our population produce children through mutation and crossover.
\subsubsection{Evaluation.}
To evaluate the networks in our population during training, we use a dataset of 100 randomly generated data-equation pairs per the constraints mentioned in sect. \ref{modelArch}. Each network is given the $XY$ data and predicts a corresponding equation $\hat{eq}$. This predicted equation is then evaluated on the $X$ data to produce $\hat{Y} = \hat{eq}(x)$. The predicted and target equations are compared syntactically using symbolic loss and then semantically using numeric loss on the $Y$ and $\hat{Y}$ values.

\subsubsection{Symbolic Loss}
Symbolic metrics compare two equations symbol-by-symbol (in our case, primitive-by-primitive) in order to determine how closely related the symbols that make up the equations are.

To determine the syntactic correctness of the predicted equations, each is evaluated symbolically by cross-entropy (CE) loss, given by 
\begin{equation}
  CE(eq,\hat{eq}) = -\frac{1}{N}\sum_{n=1}^N \sum_{c=1}^C eq_{n,c}\log \frac{\exp({\hat{eq}_{n,c}})}{\sum_{i=1}^C \exp({\hat{eq}_{n,i}})}
\end{equation}
where $eq$ and $\hat{eq}$ are the true and predicted tokenized equations respectively, N is the number of samples, and C is the number of possible tokens.
\subsubsection{Numeric loss}
Numeric metrics compare two equations by the $Y$ data produced when the equations are evaluated on some $X$ values to determine how closely related the data the equations are describing is.

To determine the semantic correctness of the predicted equations, they are evaluated on the given $X$ values, and the predicted $\hat{Y}$ values are used to calculate each network's mean squared error (MSE) loss, given by 
\begin{equation}
  \textrm{MSE}(Y,\hat{Y}) = \frac{1}{N} \sum_{i=1}^N (y_i-\hat{y_i})^2
\end{equation}
where ${Y}$ and $\hat{Y}$ are the true and predicted $Y$ values respectively and N is the number of samples.
\subsubsection{Selection.}
For our evolutionary procedure, selecting which population members survive each generation depends on how well the population members perform. Due to the necessity of $\hat{Y}$ values in order to use numeric metrics, the predicted equations must be valid equations and not just a random sequence of primitives. Thus, if the population members are unable to produce syntactically valid equations, a selection strategy that only uses symbolic loss must be used as numeric loss cannot be calculated. When numeric loss can be calculated, then selection is able to be dual-objective between numeric and symbolic loss.

To select the population members to survive, we use two different strategies and switch between them as necessary, following the precedent of modularity~\cite{Ellefsen2015,Clune2013}. This is needed due to the lack of syntactically valid equations in early generations (Fig.~\ref{fig:percdata}). In the case where there are networks in the population that are unable to produce syntactically valid equations that can be evaluated to obtain $\hat{y}$ values, all networks are sorted in order of increasing CE loss and the top 15 are chosen to survive as parents. 

Once the entire population of networks are able to produce valid equations, the selection switches to tournament Pareto-front selection. Specifically, two individuals from the population are randomly chosen, and if one individual dominates the other then that individual is chosen to survive. The other individual is removed from the population. If both symbolic and numeric loss is equal, then the younger individual is chosen to survive. If neither is dominant, both individuals remain in the population. This continues until the population size matches the chosen number of parents. Thus, the 15 networks (half the population size) that dominate the rest of the population by having lower CE and MSE loss are chosen to survive as parents.

\subsubsection{Mutation \& Crossover.}
Here, we provide details on the specifics of how parent population members are mutated and crossed-over in order to produce offspring.

For our method, we evolve a population of 30 networks, 15 parents and 15 children. We evolve for 10,000 generations, using a mutation rate and crossover rate of 50\%. Only the weights of the network are evolved, bias terms are left as-is.

The 15 children of each new generation are chosen from all parents. For each child, mutation and crossover occurs layer-wise according to the mutation and crossover rates. If a layer is chosen to be mutated, each weights in that layer is mutated by a random amount in the range of $[-0.01,0.01]$. If a layer is chosen to be crossed over, another unique parent is chosen randomly from the population. The weights of half of the corresponding layer from the first parent are then replaced with those weights from the second parent.

\subsection{Testing.}
\label{testing}
In this section, we discuss the various strategies we use to test the performance of our population of evolved SRNE networks in order to comprehensively understand both how well they compare to other GNSR methods and how well they are able to perform the task of generalizing to produce an equation for any given dataset.

To test the networks after evolution, a dataset of 100 $XY$ data-equation pairs is used. The equations are chosen from the benchmark equations recommended by Bertschinger et al.~\cite{Bertschinger2023} that contain only those mathematical operators available in our equations' primitive set: 
\begin{equation}
    x^3+x^2+x
\end{equation}
\begin{equation}
    x^4+x^3+x^2+x
\end{equation}
\begin{equation}
    \sin{x} + \sin{(x+x^2)}
\end{equation}
\begin{equation}\label{eq:excluded}
    \sin(x*e^x)
\end{equation}
\begin{equation}
    x + \log (x^4)
\end{equation} 
The testing dataset contains 20 samples of each equation with a set of 30 randomly generated $X$ values and corresponding $Y$ values.

To report symbolic testing accuracy, we use CE for the Pareto-fronts formed from the final generation's surviving population over 20 trials. 

Instead of using MSE for testing accuracy, we use normalized mean squared error (NMSE), given by
\begin{equation}
  \textrm{NMSE}(Y,\hat{Y}) = \frac{1}{n} \sum_{i=1}^n \frac{(y_i-\hat{y_i})^2}{\lVert y_i+\epsilon\rVert_2}
\end{equation}
 where ${Y}$ and $\hat{Y}$ are the true and predicted $Y$ values respectively, n is the number of samples, and $\epsilon$ is a small term added to ensure no divide-by-0 errors occur. NMSE is a metric used in other GNSR literature, so we use this for testing accuracy for ease of comparison.
 
In addition to reporting the Pareto-fronts from testing, we also test the SRNE networks on equations that were unseen by the network during pre-training or evolution. The four equations that make up this unseen testing set are the same as the testing equations with the exception of equation \ref{eq:excluded}, which happened to be present in the SRNE networks' pre-training. These unseen equations represent equations that the networks' weights were not trained to produce during pre-training or evolution, in order to show that our method can predict equations outside of its training set. SRNE networks' performance is reported on 4 metrics for variety in performance results. For symbolic metrics, CE and tree edit distance (TED) is reported. To calculate TED, first both the target and predicted equations are simplified to combine any duplicate terms and to ensure that both equations are ordered the same way. The equations are then transformed into equation trees where nodes represent the primitives that make up the equation and edges form the structure of the equation.  Then the number of changes that needed to be made to transform the predicted equation tree to the target equation tree is calculated. This number of changes is then reported as the TED. For numeric metrics, NMSE and the coefficient of determination ($R^2$) is reported. $R^2$ is calculated by 
\begin{equation}
  R^2(Y,\hat{Y}) = 1 - \frac{\sum_{i=1}^N (y_i - \hat{y_i})^2}{\sum_{i=1}^n (y_i - \bar{y})^2}
\end{equation}
where ${Y}$ and $\hat{Y}$ are the true and predicted $Y$ values respectively, $\bar{y}$ is the mean of $Y$ values, and N is the number of samples. In order to treat $R^2$ as loss, we report $1-R^2$.

\subsection{Comparison to state-of-art.}
\label{comparison}
We compare against three GNSR methods from literature: Neural Symbolic Regression that Scales (NSRTS)~\cite{Biggio2021}, SymbolicGPT (SGPT)~\cite{Valipour2021}, and Symformer~\cite{Vastl2022}. NSRTS is a transformer network that is trained on millions of data-equation pairs generated by ~\cite{Biggio2021}'s method and CE loss. SGPT is a two-network model: a convolutional PointCloud network to produce an order-invariant data representation and a transformer that predicts equations. It is trained on ten thousand data-equation pairs generated with ~\cite{Valipour2021}'s equation generation method and CE loss. Symformer is a transformer architecture that predicts both equations and coefficients, trained on millions of data-equation pairs of ~\cite{Vastl2022}'s equation generation method. It uses CE loss for predicted equations and MSE loss for predicted coefficients. For all methods, we use pretrained networks provided publicly by the authors.

To obtain results from these pretrained networks, we create datasets containing 100 of our testing equations and randomly generated $X$ values, to the specification of each method (in regards to range of $X$ values, number of $X$ values, etc.). We then run each respective pretrained network on its' dataset. For the numeric metrics, we take the median value over the 100 predicted equations due to $NMSE$ and $R^2$ both having unbounded limits for poor results while perfect results are bounded by 0 and 1 respectively. For the symbolic metrics, we take mean value over the 100 predicted equations.

\section{Results}
Here we discuss the results of the dual-objective evolution of SRNE networks. For all results, we ran 20 independent trials wherein a population of neural networks is evolved to the above specifications. The only commonality between trials are the 25 pretrained synaptic weights that the initial generation's parents pull from as specified in sec. \ref{pre-training}. 
We first show the fitness over time (sect. \ref{fot}), then Pareto-fronts for evolution and testing, (sect. \ref{Pareto-front}) and finally the performance on unseen equations  (sect. \ref{unseen-eq}).
\subsection{Fitness Over Time}
\label{fot}
\begin{figure}[t]
  \centering
  \includegraphics[width=.9\linewidth]{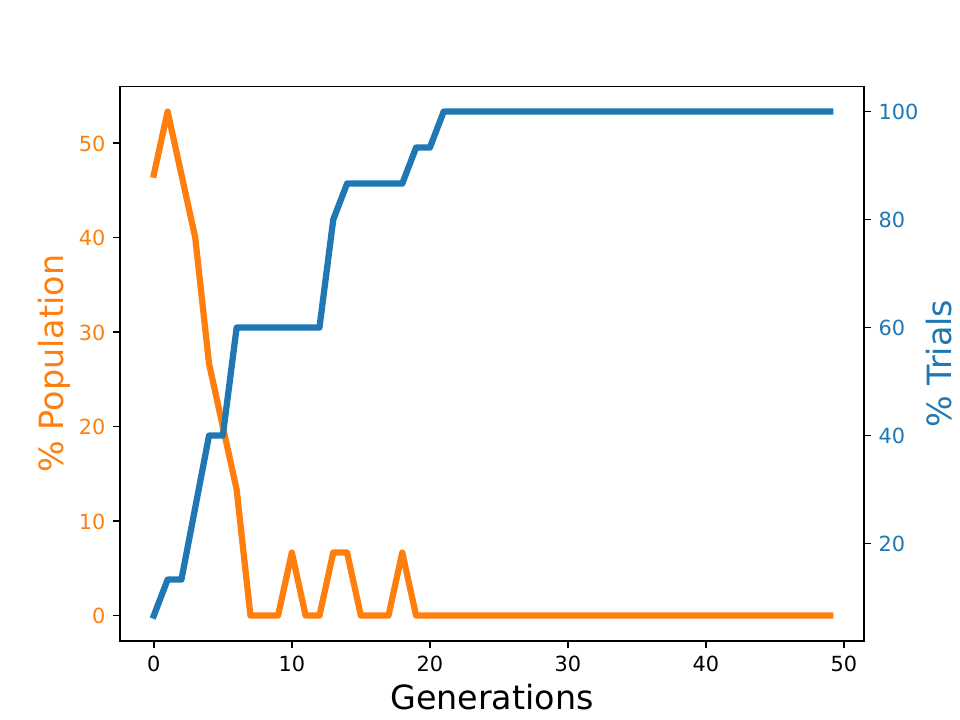}
  
  \caption{Emergence of valid equations. 
  Blue is the percentage of the 20 trials that have started producing valid equations, enabling dual-objective optimization. 
  Orange is the percentage of proportion of one trial that are producing invalid equations over the generations. Only 50 generations are shown for readability.}
  \label{fig:percdata}
  \vspace{-0.2cm}
\end{figure}
\begin{figure}[t]
  \centering
  \includegraphics[width=.9\linewidth]{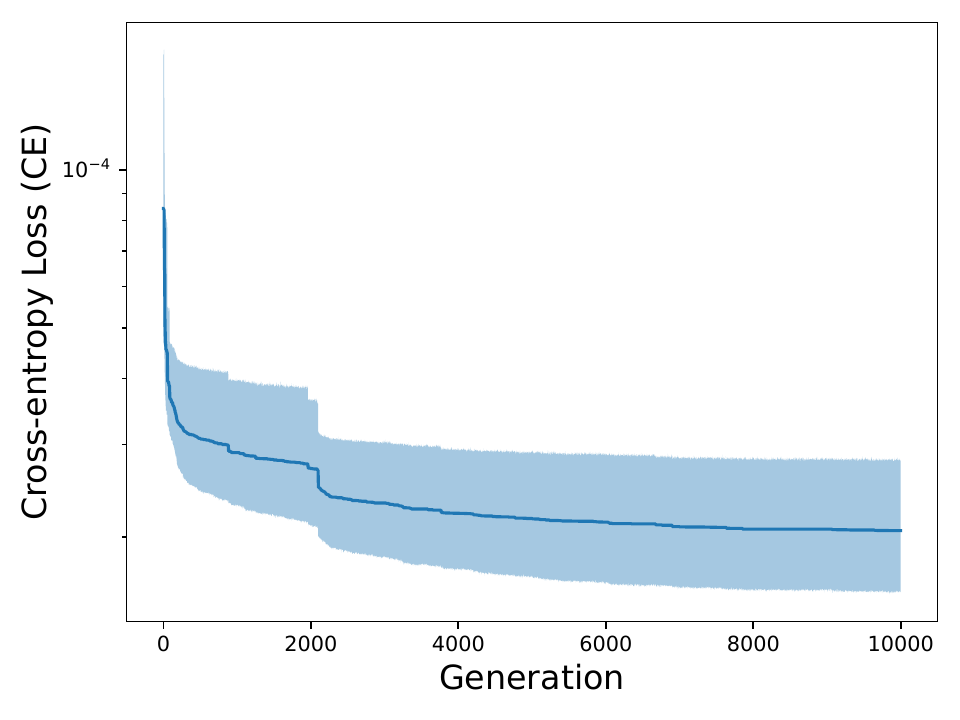}
  \vspace{-0.2cm}
  \caption{Average symbolic loss over 20 trials. Horizontal axis shows generations and the vertical axis is cross-entropy (CE) loss. A 95\% confidence interval is shown. Vertical axis is logarithmic for readability.}
  \label{fig:symfitness}
  \vspace{-0.1cm}
\end{figure}

\begin{figure}[t]
  \centering
  \includegraphics[width=.9\linewidth]{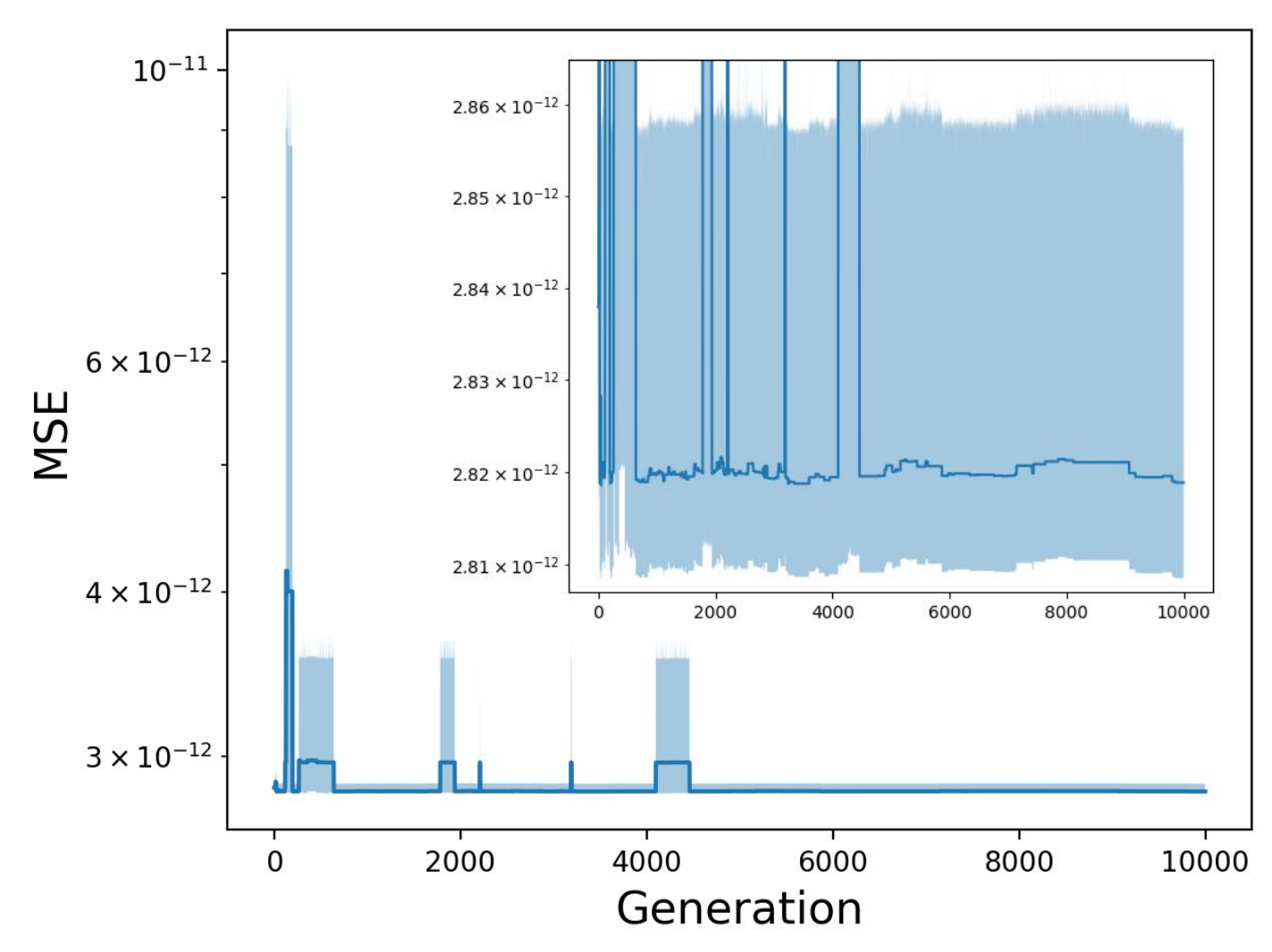}
  \vspace{-0.2cm}
  \caption{Average numeric loss over 20 trials. Horizontal axis shows generations and the vertical axis is mean squared error (MSE) loss. A 95\% confidence interval is shown. Vertical axis is logarithmic for readability on the outer plot, but is linear on the inset plot.}
  \label{fig:numfitness}
  \vspace{-0.3cm}
\end{figure}
Figure \ref{fig:symfitness} shows the fitness of the best performing population member by CE loss for each generation, averaged over 20 trials. At generation 0, the average CE loss is 8.44e-05 and by the final generation, the average CE loss is 2.06e-05. Across the trials, there is a significant difference between generation 0 and generation 9999 ($p\ll0.05$), with much of the improvement occurring in the first 500 generations.

Figure \ref{fig:numfitness} shows the numeric fitness of the best performing population member for each generation, averaged over 20 trials. As discussed in sec. \ref{evolution}, the selection of best performing population member by numeric fitness metrics is dependent on the population's ability to predict syntactically valid equations. Figure \ref{fig:percdata} blue curve shows the percentage of the population over the first 50 generations that are producing all valid equations and are therefore able to use multi-objective optimization. Over our trials, it took a minimum of 0, maximum of 21, and average of 8.5 generations to be able to use multi-objective selection and optimize for numeric fitness at all. Figure \ref{fig:percdata} orange curve shows the percentage of the population that is predicting invalid equations over the first 50 generations for 1 trial. For this trial, at generation 0 47\% of the population is predicting invalid equations, and it takes until generation 19 for the population to be consistently producing all valid equations. Across all trials, the first 21 generations had an average numeric loss of 2.829e-12. At the final generation, there was an average numeric loss of 2.819e-12. Across the 20 trials, there is a significant difference between the first 21 generations and generation 9999 ($p<0.05$).

\subsection{Pareto Fronts}
\label{Pareto-front}
\begin{figure*}[t]
  \centering
  \includegraphics[width=.9\textwidth]{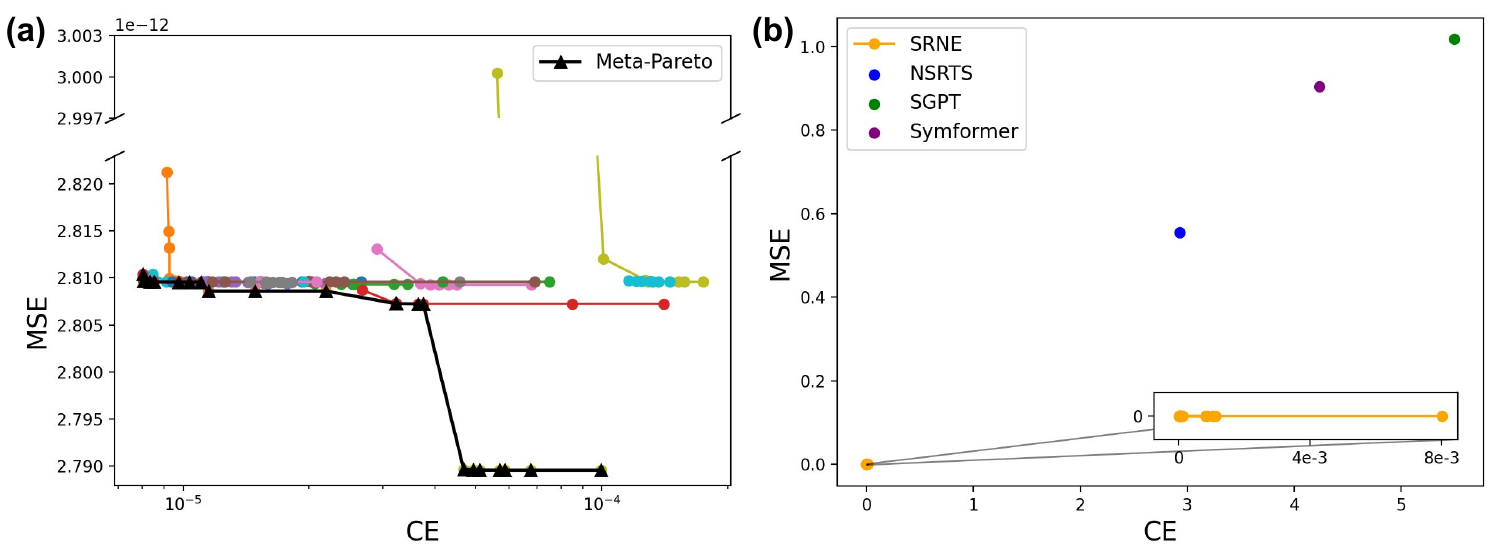}
  \vspace{-0.2cm}
  \caption{(a): Pareto-fronts for the population of evolved networks at the final generation over 20 trials, evaluated on the training equations. Black is the meta-Pareto drawn from the combination of all trials' populations. The horizontal axis is logarithmic for readability. (b): Performance of other GNSR methods from literature compared to SRNE (ours) networks on unseen test equations. Compared methods are averaged over the same 100 test equations. Zoomed window (lower right) shows the meta Pareto-front of the SNRE networks, assembled from 15 final evolved networks from 20 trials that are evaluated after evolution on the testing data-equation pairs.}
  \label{fig:Pareto}
  \vspace{-0.2cm}
\end{figure*}
Figure \ref{fig:Pareto} shows the Pareto-fronts from the SRNE evolved networks. Fig. \ref{fig:Pareto}(a) shows the 20 Pareto-fronts formed by the final population. The black line is the meta-Pareto-front formed by combining populations from all 20 trials and then obtaining the Pareto-front for this meta-population. Overall, the lowest symbolic loss achieved in the meta-front is a CE of 8.02e-6 and the lowest numeric loss is an MSE of 2.79e-12.

Fig. \ref{fig:Pareto}(b) shows the performance of SRNE networks as compared to other GNSR networks. As can be seen from the plot, the SRNE Pareto-front is able to dominate the results of the other networks. Across the networks in the Pareto front, the numeric (NMSE) loss is consistently 0.0, leading to a flat pareto-front. The symbolic (CE) loss ranges from 1.13e-5 to 0.008. 
\subsection{Unseen Equations}
\label{unseen-eq}
\begin{table}[t]
\center{
\caption{\label{table:unseen} Symbolic and numeric loss of the SRNE networks compared with state-of-the-art GNSR methods from the literature. Comparisons are conducted on a random sampling of 100 instances of the first 4 test equations. Mann-Whitney U tests for significance were performed within each column with a Bonferroni correction applied for the pairwise comparisons. * indicates that SRNE performed significantly better ($p\ll0.001$) than the indicated GNSR method, for that metric.}
\vspace{-0.2cm}
\begin{tabular}{| l || c | c | c | c | } \hline

        & \multicolumn{2}{c|}{Symbolic loss} & \multicolumn{2}{c|}{Numeric loss} \\\hline        
Network & CE & TED & NMSE &$1-R^2$  \\ \hline \hline

SGPT  (ref.~\cite{Valipour2021})   & 5.496* & 16.039* & 1.217* & 0.955*     \\ \hline
Symformer (ref.~\cite{Vastl2022}) & 4.121* & 12.460* & 0.444* & 0.049*     \\ \hline
NSRTS (ref.~\cite{Biggio2021})   &  3.508* & 6.430* & 0.963* & 0.105*     \\ \hline
\textbf{SRNE (ours)}                     & \textbf{3.17e-5} & \textbf{0} & \textbf{0.0} & \textbf{0.0}    \\ \hline
\end{tabular}
}
\vspace{-0.3cm}
\end{table}
Table \ref{table:unseen} shows the performance of a randomly chosen SRNE network on previously unseen equations as compared with other GNSR methods from the literature, details of which are discussed in sec. \ref{comparison}. As the other methods are all pretrained models that we simply tested on the unseen equations, we cannot guarantee that their GNSR methods did not see these equations during training. Across all symbolic and numeric metrics, our method performs significantly better than the other methods from literature.

\section{Discussion}
\begin{figure}[t]
  \centering
  \includegraphics[width=0.8\linewidth]{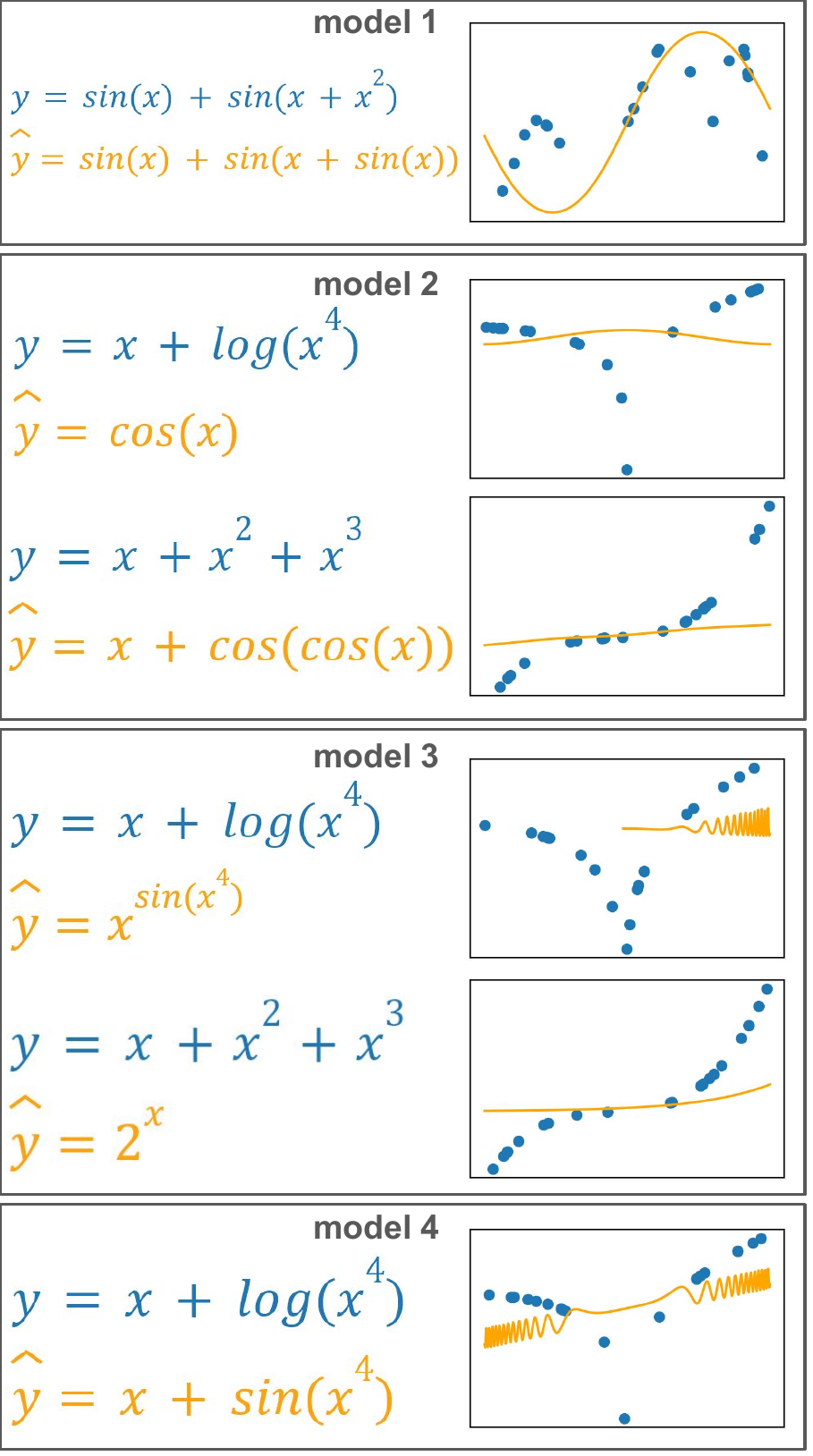}
  \caption{Examples of various evolved SRNE networks and cases where said networks were unable to perfectly predict the target equation. The equations show the target (blue) and predicted (orange) equations. The plots show the target $Y$ data (blue) and the predicted $\hat{Y}$ (orange).}
  \label{fig:eqviz}
  \vspace{-0.3cm}
\end{figure}
\begin{table}[t]
\center{
\caption{\label{table:gpsr} Training and testing inference time as well as numeric and symbolic loss of a GPSR method (PySR) compared to our SRNE method. Training and testing times are the wall time of that stage of the method, given in seconds. Both SRNE and PySR results are averaged over 20 independent trials. Mann-Whitney U tests for significance were performed within each column with a Bonferroni correction applied for the pairwise comparisons. * indicates that SRNE performed significantly better ($p\ll0.001$) than PySR for that metric.}
\vspace{-0.2cm}
\begin{tabular}{| l || c | c | } \hline
       
   & PySR (ref.~\cite{Cranmer2023}) & SRNE  \\ \hline \hline
     \multicolumn{3}{|c|}{Wall Time} \\ \hline
Pre-training (sec. \ref{pre-training})          & N/A & 10412.50 s      \\ \hline
Evolution (sec. \ref{evolution})          & N/A &  64644.87 s    \\ \hline
Total Training          & N/A & 75057.37 s     \\ \hline
Testing (sec. \ref{testing}) & 25892.41 s & 3.76 s      \\ \hline\hline
\multicolumn{3}{|c|}{Testing Loss} \\ \hline
TED           & 10.435 & \textbf{0.122*}     \\ \hline
MSE           & 2.920 & \textbf{0.083*}     \\ \hline
\end{tabular}
}
\vspace{-0.3cm}
\end{table}
\begin{figure}[t]
  \centering
  \includegraphics[width=0.8\linewidth]{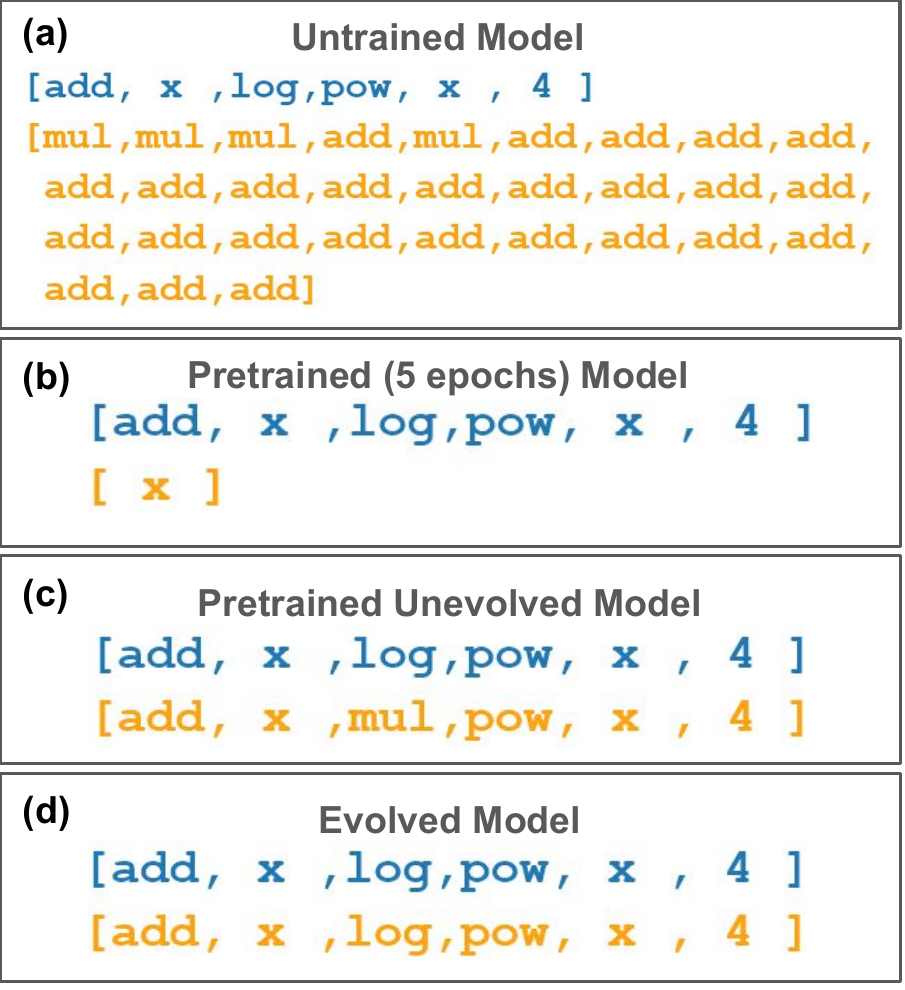}
  \caption{The progression of how the SRNE networks produce increasingly more accurate equations. Test target and predicted equations for a (a) completely untrained network, (b) network pretrained on gradient descent CE for 5 epochs, (c) fully pretrained but unevolved network, and (d) fully evolved network.}
  \label{fig:eqprog}
  \vspace{-0.4cm}
\end{figure}
Figure \ref{fig:symfitness} shows that, although the improvements tend to be small and can take many generations, over the course of the entire evolution, symbolic loss is able to significantly improve over the initial starting population of networks pretrained using purely single-objective backpropagation. Figure \ref{fig:numfitness} similarly shows that over the 10000 generations, there is a significant reduction in numeric loss. The spikes in the numeric fitness seen in Fig. \ref{fig:numfitness} can be explained by our selection process: if the networks are not predicting equations capable of being evaluated on numeric metrics, then the selection is solely on symbolic fitness. Thus, there are some generations where the numeric fitness temporarily worsens before improving. As evolution continues, this behavior disappears as the networks are capable of predicting valid equations and the optimization is more consistently decreasing. Evidence of the increasing stability of prediction of valid equations by the population is seen in Figure \ref{fig:percdata}. The blue curve showing the percent of trials that are able to start multi-objective optimization due to invalid equations increases steadily over the first 21 generations. However, as seen in the orange curve, even after all population members are able to predict valid equations for one generation, there may be future instability in children that cannot produce valid equations. This curve also reveals that even when a child with this behavior occurs, it is quickly eliminated from the surviving population. Thus, the network is able to achieve significantly lower loss by later generations.

Figure \ref{fig:Pareto}(a) shows that across all 20 trials, there is relative consistency in the evolution, as all of the final Pareto fronts tend to have approximately the same loss values. Overall, the networks are able to fulfil the dual purpose of minimizing both symbolic and numeric loss, reducing both to near zero by the end of evolution. Fig. \ref{fig:Pareto}(b) further shows the performance of the SRNE method: the results obtained from the dual-objective optimization are able to dominate the results of other GNSR networks on the same testing equations. In fact, the SRNE method achieves perfect numeric loss (NMSE of 0) across its dominant Pareto-fronts. It also achieves near perfect symbolic loss with a CE loss of near zero. The inability of the SRNE networks to achieve a perfect CE while being able to get a zero MSE can be explained by the fact that for some mathematical expressions, tokenization order does not matter for a numeric metric but does for a symbolic metric. 

As Fig. \ref{fig:Pareto}(b) shows only the Pareto-front which performs perfectly across numeric loss, we included Fig. \ref{fig:eqviz} as a few examples of SRNE networks that are not able to perfectly predict all 100 test equations. As can be seen, some networks are not able to perfectly predict all of the correct operators in the test equations. Model 4 shows a common case wherein $\log(x)$ in the target equation is predicted to be $\sin(x)$. Overall, though, the SRNE networks are able to predict equations more accurately than other GNSR methods both symbolically and numerically.

Table \ref{table:gpsr} shows the results of SRNE compared with a standard GPSR method, PySR~\cite{Cranmer2023}. As seen in the comparison of numeric and symbolic losses, our method performs significantly better across both. This is in spite of the fact that PySR (as with all GPSR methods) is evolved per inference equation. By comparison, our method is trained to output any inference equation (Fig. \ref{fig:teaser}(a)). This is seen in the comparisons of training time, where PySR has no training time on a large corpus, while ours has quite a large training wall-time of approximately 21 hours. This represents the time to pretrain 25 networks for 50 epochs on 5000 equations and then evolve the population of networks for 10,000 generations on 100 equations. However, when it comes to inference time of the 125 testing equations, it only takes our method approximately 3.76 seconds as compared to the approximately 7 hours of wall time that PySR takes. This means that our method spends on average 0.03 seconds per equation while PySR takes on average 207.14 seconds per equation. While SRNE's initial training period is significant, as PySR scales linearly per equation and assuming that times hold with an increased number of evaluated equations, SRNE becomes more computationally efficient than PySR after 363 equations.

Figure \ref{fig:eqprog} further emphasizes the importance of the dual-objective training by showing the progression of how increasingly more pretrained and evolved networks predict more accurate equations. Fig. \ref{fig:eqprog}(a) shows how a completely untrained network predicts a syntactically invalid expression. After a short amount of initial pretraining, Fig. \ref{fig:eqprog}(b) that the network has learned to predict a shorter "equation", but not a particularly accurate one. Then, Fig. \ref{fig:eqprog}(c) shows a fully pretrained network that can predict a mostly symbolically accurate equation. However, this equation is not a valid equation as the primitives are not in the correct order. Thus, though the equation is more symbolically accurate, it is numerically inaccurate. Finally, Fig. \ref{fig:eqprog}(d) shows an equation produced by an evolved SRNE network that can predict a valid equation that is accurate both symbolically and numerically. This highlights not just the importance of multi-objective optimization for GNSR, but also how when backpropagation and evolution are combined, their advantages are jointly leveraged to produce the most accurate result.

Taken together, Figs. \ref{fig:symfitness}\&\ref{fig:numfitness} empirically prove that the dual-objective optimization of symbolic and numeric loss performs better at the task of training networks to produce accurate equations than state-of-the-art approaches. Although symbolic and numeric loss are related (as perfectly symbolically accurate equations will by definition produce perfectly numerically accurate equations), it has been shown that a target and predicted equation can be close semantically but not syntactically, and vice versa~\cite{Bertschinger2023}. We have thus proved through our results the necessity of the dual-objective training on GNSR networks.

Table \ref{table:unseen} shows that our network can generalize to unseen equations, achieving perfect results on the numeric metrics and near-perfect results on the symbolic metrics. Importantly, our method achieves significantly better results across the unseen equations than other GNSR methods do on these same equations. This is in spite of the fact that it is possible that those methods did see these equations during training. We cannot guarantee one way or another if the testing equations used in the unseen results were included in the training of other GNSR results, as we ran these results on pretrained networks and did not have access to the training datasets to confirm the absence of the equations. It is all the more impactful, then, that our SRNE method was able to outperform the other methods across all testing metrics on these unseen equations. Our network is not only better at predicting \textit{known} equations, but can truly fulfil the purpose of symbolic regression: to be able to generalize to \textit{unknown} equations given only the data that those equations describe without interference from a human expert.

\section{Conclusion}
We have shown empirically that networks that are trained dual-objectively on symbolic and numeric losses dominate the results of single-objective trained networks. The key difference in our SRNE method is this dual-objective optimization, as otherwise our networks' architecture is similar to existing GNSR architectures. 

We have also shown the power of combining evolutionary algorithms with gradient descent for the purpose of symbolic regression. To the best of our knowledge, to date, there is no published GNSR gradient descent method that is able to optimize on numeric loss. This is made possible through the use of evolutionary algorithms, and the evolution of our SRNE networks was shown to significantly improve the accuracy of the predicted equations. We were able to achieve significantly better results after evolution than the initial population of gradient-descent pretrained networks. We hope that our method serves as an example of how evolutionary algorithms can be combined with gradient descent for deep neural networks in order to leverage the advantages that both methodologies provide. We also hope that it serves as proof of the necessity of dual-objective training for GNSR with the purpose of having predicted equations that are accurate in both form and function.

In future work, we wish to continue this research into both dual-objective optimization in GNSR and the combination of evolutionary algorithms with gradient-descent training. Our method leveraged the power of evolutionary optimization of synaptic weights, but evolutionary algorithms can also help optimize network architecture~\cite{Stanley2002}. Through employing an algorithm such as NEAT~\cite{Stanley2002}, we may be able to achieve further success in predicting equations that are more accurate both syntactically and semantically by exploring hyperparameter and architecture optimization of the GNSR networks. Our method explored only optimizing synaptic weights, with the structure of the population networks held constant. It is possible that allowing for evolution of not just the weights but the structure of the network would allow for more efficient or better performing networks.
\begin{acks}
The authors wish to acknowledge support from Army Research Office contract \#W911NF2310100. Computations were performed on the Vermont Advanced Computing Center supported in part by NSF award No. OAC-1827314.
\end{acks}
\bibliographystyle{ACM-Reference-Format}
\bibliography{references}

\appendix

\end{document}